\documentclass[10pt,twocolumn,letterpaper]{article}

\usepackage{cvpr}              

\usepackage{graphicx}                    

\usepackage{booktabs}
\usepackage{xcolor}
\definecolor{bggray}{RGB}{235, 235, 233}
\definecolor{prompt}{RGB}{245, 85, 85}
\definecolor{sent}{RGB}{83, 142, 33}

\definecolor{verylightgray}{rgb}{0.95, 0.95, 0.95}
\definecolor{verylightgreen}{rgb}{0.9, 1, 0.9}
\definecolor{verylightred}{rgb}{1, 0.9, 0.9}

\usepackage{xcolor}
\definecolor{mylightgray}{gray}{0.5}
\definecolor{mydarkgreen}{RGB}{0, 150, 0}
\definecolor{mydarkred}{RGB}{200, 0, 0}
\definecolor{myblue}{RGB}{0, 0, 255}
\definecolor{myorange}{RGB}{255, 100, 0}

\usepackage{xcolor}
\usepackage{color, colortbl}
\definecolor{verylightgreen}{rgb}{0.9, 1, 0.9}
\definecolor{verylightred}{rgb}{1, 0.9, 0.9}
\definecolor{verylightblue}{rgb}{0.8, 0.9, 1.0}
\definecolor{verylightorange}{rgb}{1.0, 0.8, 0.6}
\definecolor{verylightgray}{rgb}{0.95, 0.95, 0.95}

\usepackage{subcaption}
\usepackage[margin=1in]{geometry}

\usepackage{amsmath}

\usepackage{algorithm}
\usepackage{algpseudocode}
\newcommand{\comment}[1]{}

\definecolor{cvprblue}{rgb}{0.21,0.49,0.74}
\usepackage[pagebackref,breaklinks,colorlinks,allcolors=cvprblue]{hyperref}

\def\paperID{15} 
\def\confName{CVPR}
\def\confYear{2025}

\title{TRAVEL: Training-Free Retrieval and Alignment for Vision-and-Language Navigation}

\author{Navid Rajabi\\
George Mason University\\
{\tt\small nrajabi@gmu.edu}
\and
Jana Ko{\v{s}}eck{\'a}\\
George Mason University\\
{\tt\small kosecka@gmu.edu}
}

\begin{document}
\maketitle

\begin{abstract}
In this work, we propose a modular approach for the Vision-Language Navigation (VLN) task by decomposing the problem into four sub-modules that use state-of-the-art Large Language Models (LLMs) and Vision-Language Models (VLMs) in a zero-shot setting. Given navigation instruction in natural language, we first prompt LLM to extract the landmarks and the order in which they are visited. Assuming the known model of the environment, we retrieve the top-k locations of the last landmark and generate $k$ path hypotheses from the starting location to the last landmark using the shortest path algorithm on the topological map of the environment. Each path hypothesis is represented by a sequence of panoramas. We then use dynamic programming to compute the alignment score between the sequence of panoramas and the sequence of landmark names, which match scores obtained from VLM.  Finally, we compute the nDTW metric between the hypothesis that yields the highest alignment score to evaluate the path fidelity. We demonstrate superior performance compared to other approaches that use joint semantic maps like VLMaps \cite{vlmaps} on the complex R2R-Habitat \cite{r2r} instruction dataset and quantify in detail the effect of visual grounding on navigation performance.

\end{abstract}

\section{Introduction}
Vision-and-Language Navigation (VLN) task 
involves controlling an agent, either in simulation or in the physical world, to navigate through an environment by following natural language instructions. Consider an example in Fig. 1 where the agent is required to 
follow the instructions in a specific environment. This task requires parsing the language input (e.g., “Turn left in the hallway, go to the kitchen, and stop by the sink”), grounding the phrases to visual concepts such as scenes, landmarks, and actions (e.g., turn left) as well temporal cues (e.g., turn before).

\begin{figure}[!h]
  \centering
  \includegraphics[width=\linewidth]{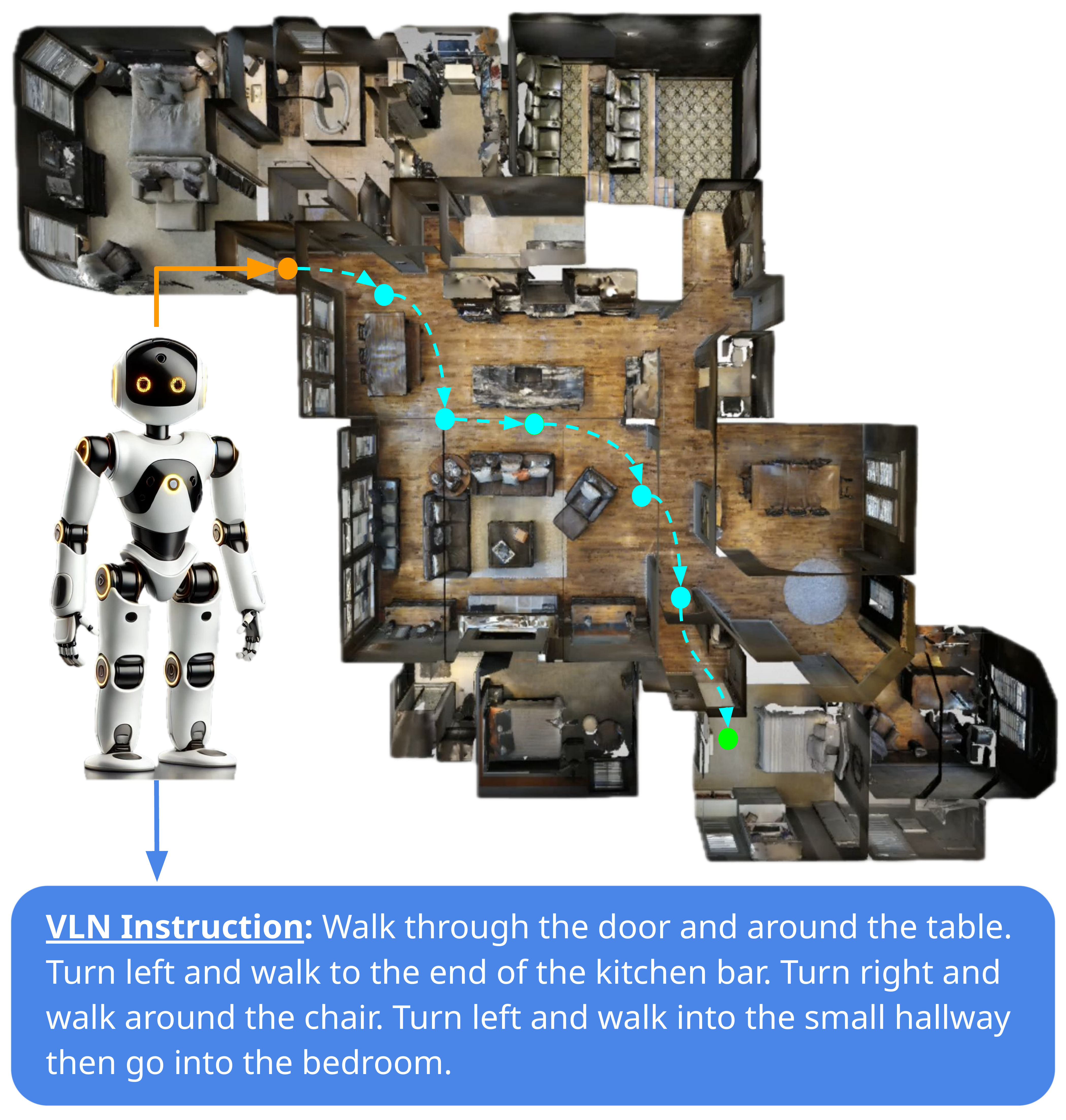}
  \caption{Bird's Eye View visualization of a sample VLN episode from R2R dataset~\cite{r2r}.}
  \label{fig:vlnsampleepisode}
\end{figure}



One class of approaches formulates the Vision-Language Navigation task as a supervised multi-modal sequence-to-sequence learning task, where the learner is given episodes of natural language instructions, along with visual observations and navigation actions. These approaches were supported by large-scale datasets of navigation instructions, e.g., Room-2-Room (R2R) \cite{r2r}, in Matterport3D \cite{chang2017matterport3d} indoor environments, providing the agent with panoramic images from different locations. The sequence-to-sequence methods varied in their multi-modal language and vision architectures, training techniques, and choices of representations, gradually improving the benchmark performance.  Despite these improvements, non-negligible gaps still exist between machines' and human performance on existing benchmarks, the performance suffers in novel environments and in the presence of more complex variation of instructions.

Another class of methods pursued a more modular approach, using 
or learning separate modules for processing natural language inputs, using semantic segmentation or detection to ground noun phrases images and in the map and integrating these with more traditional map-based navigation. These methods, however, use simple natural language instructions and are typically evaluated on small-scale datasets
\cite{liu2023lang2ltl, vlmaps}.

\noindent
{\bf Contributions.}
In the presented work, we pursue a modular approach, 
where we exploit zero-shot capabilities of the state-of-the-art LLMs for understanding and parsing navigation instructions and VLMs for grounding landmark names in the visual observations. The navigation component is 
carried out by finding a path in the topological map of the environment that is best 
aligned with the navigation instructions.
The map is acquired using the training 
episodes from R2R dataset~\cite{r2r}, and the alignment score is computed using dynamic programming, where the costs of individual 
steps are obtained from the state-of-the-art Vision-Language Model. The presented modular approach demonstrates superior performance over occupancy map-based approaches and reveals current strengths and weaknesses of the state-of-the-art LLMs and VLMs for vision-language instruction following.

\section{Related Work}
For the purpose of our exposition, the existing works on Vision Language Navigation can be partitioned into end-to-end and modular approaches. The end-to-end methods take the natural language instructions, visual observations, and actions and train a multi-modal sequence-to-sequence model, and in the inference stage, given the instruction and initial view, the model generates the sequence of actions while ingesting additional views.  
The modular approaches integrate LLMs, VLMs, or both with more traditional map-based representations along with a common robotics navigation stack comprised of basic navigation skills that are not learned.  \\
\noindent
\textbf{End-to-end approaches.} These methods typically adopt a sequence-to-sequence model, taking as an input the language instruction and visual information and outputs the sequence of low-level navigation actions (move, turn left/right) or local waypoints. During the forward pass, the entire instruction is processed by the Language Encoder (e.g., LSTM/transformer). The aggregation of the context vectors, plus the encoded current view of the agent, is then fed to the Action Decoder (e.g., LSTM/transformer) that generates the next \texttt{action}. The decoder continues to predict actions until it generates the \texttt{STOP} action. 
The mixture of Reinforcement Learning (RL) and Imitation Learning (IL) has been commonly used for training these 
models~\cite{tan2019envdrop}. 
The existing approaches proposed different variations of model architectures, training strategies and choice of representations~\cite{vln, tan2019envdrop, fried2018speaker, wang2019reinforcedxmm, vlnce, hong2021vlnrecbert, moudgil2021soat, chen2021historyhamt, georgakis2022crosscm2}  typically using the Room-to-Room (R2R)~\cite{r2r} and Room-Across-Room (RxR)~\cite{rxr} benchmarks for training and evaluation. The natural language instructions in these benchmarks are quite complex, with an average length of $\sim$ 26 words. 
These approaches have made substantial improvements in past years, mostly thanks to increasing the number of training episodes and auxiliary tasks that support grounding \cite{wang2022lessismore} and instruction generation \cite{fried2018speaker, kamath2023marval}. It has been shown \cite{zhu2021diagnosing} that the performance of the existing methods continues to be severely compromised by the inability to ground landmarks, understand spatial relationships, as well as grounding of action phrases. The ability to ground landmarks is more critical for indoor environments, while in outdoor settings, the grounding of actions in navigation instructions is more critical.
Furthermore, RL \& IL require a large number of high-quality training episodes, in addition to the extra computational complexity of RL due to the online interaction of the agent with the simulator/environment that makes it more difficult to scale the training \cite{kamath2023marval}.\\
%
\begin{figure*}[!h]
  \centering
  \includegraphics[width=\linewidth]{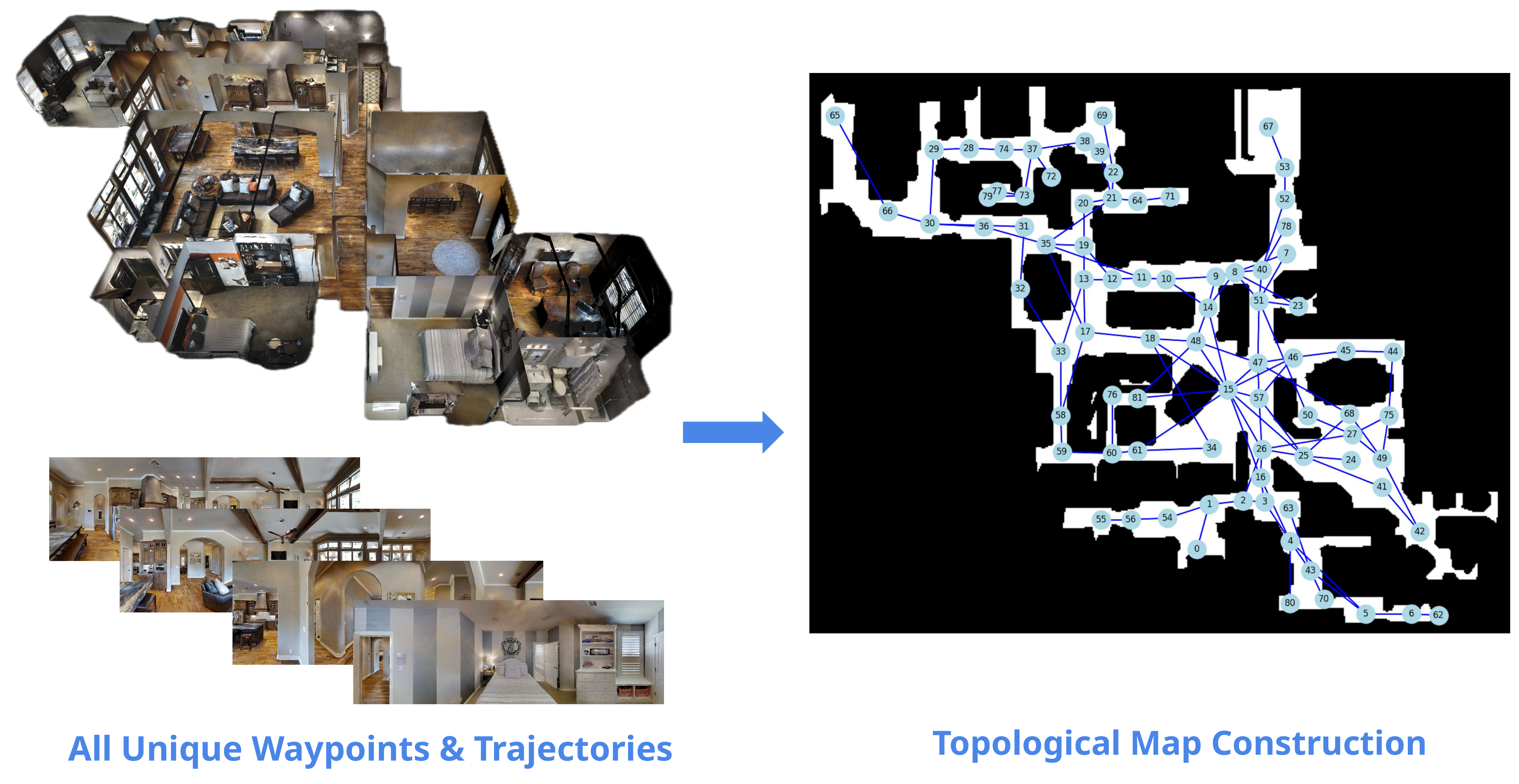}
  \caption{Topological Map Construction}
  \label{fig:vlntopo}
\end{figure*}
\noindent
\textbf{LLM and VLM based modular approaches.} Language Models were used in the past as zero-shot planners, where  \cite{huang2022lmzeroshotplanners} introduced the idea of utilizing the knowledge learned by LLMs, like OpenAI GPT-3~\cite{brown2020gpt3} and Codex~\cite{chen2021codex}, for decomposing high-level tasks (e.g. "make breakfast") to sequences of lower level skills executable by the agent. For navigation tasks, CLIP-Nav~\cite{clip-nav} utilized CLIP VLMs~\cite{clip} for grounding instruction phrases and GPT-3~\cite{brown2020gpt3} for decomposition of complex natural language instructions into phrases. In CLIP-Nav, the language instruction is decomposed using GPT-3~\cite{brown2020gpt3}, and then each sub-instruction, along with a panorama comprised of four egocentric views, is ranked by CLIP~\cite{clip} to determine the closest heading direction. The major limitations of CLIP-Nav are the dependency on the existence of a navigable graph of the environment and the poor ability of CLIP to associate landmarks with images. 
Another decomposition of the navigation task was adopted by the VLMaps~\cite{vlmaps} approach, which first builds a global joint vision-language semantic occupancy map by exploring the environment.
The cells of the map are populated by LSeg/CLIP embeddings~\cite{lseg, clip}, projected onto the grid from images. 
The navigation instructions are simpler, often resorting to point and object goal navigation, which are further translated into robotic navigation skills in the form of executable code. 

\noindent Lang2LTL \cite{liu2023lang2ltl} represents another line of work that has been proposed to use LLMs to translate free-form natural language instructions into linear temporal logic (LTL). Lang2LTL is advantageous because it disambiguates the goal specification and facilitates incorporating temporal constraints. The limitations of Lang2LTL are the need for a parallel dataset of natural language instructions and their corresponding fixed set of LTL formulas for fine-tuning the LLMs for the translation stage and the limited level of complexity of the instructions, compared to R2R \cite{r2r} and RxR \cite{rxr} benchmarks. Authors in LM-Nav~\cite{shah2023lmnav} propose a zero-shot approach for outdoor instruction following. They utilize a visual navigation system called ViNG~\cite{shah2021ving}, to construct a topological map $G$ from a set of observations, followed by extraction of landmarks $L$ from the free-form navigation instruction using GPT-3. 
CLIP is then used to infer a joint probability distribution over the nodes in $G$ and landmarks in $L$, followed by a graph search algorithm to find the optimal path that is executed by local navigation policy. The approach in LM-Nav can only navigate to a sequence of unique landmarks by design, discarding complexities like spatial clauses and fine-grained grounding
of landmark and action phrases. 


\section{Our Approach}
We introduce a modular approach for solving the VLN task using the pre-trained state-of-the-art language and vision and language models in a zero-shot setting, focusing on complex instructions in the R2R-Habitat dataset \cite{r2r}.
The agent first builds a topological map of the environment using the train split episodes of the dataset. We used all the available unique waypoints and trajectories of the environment to build the graph $G$, where each node $v$ is represented by a 360\textdegree{} RGB panorama and each edge $e$ has a weight of 1, representing the connectivity between each pair of nodes, as shown in Figure \ref{fig:vlntopo}. In this way, we ensure consistency in our evaluation process as every node of the ground-truth waypoints from the training episodes has a corresponding node in the topological map.




Then, we extract the sequence of landmarks from the natural language instruction using a pre-trained LLM, \texttt{LLama-3.1-8B-Instruct} in our case. We identify the last landmark phrase and search panoramas for the top-k most likely goal nodes. Suppose that the last landmark is \textit{bedroom}, we can locate the goals by recognizing whether the \textit{bedroom} can be found in the panoramic images associated with the graph nodes. In this way, we will narrow down the set of possible paths that lead to the goal locations. 

\begin{figure*}[!h]
    \centering
    \begin{subfigure}[t]{0.49\textwidth}
        \centering
        \includegraphics[width=\linewidth]{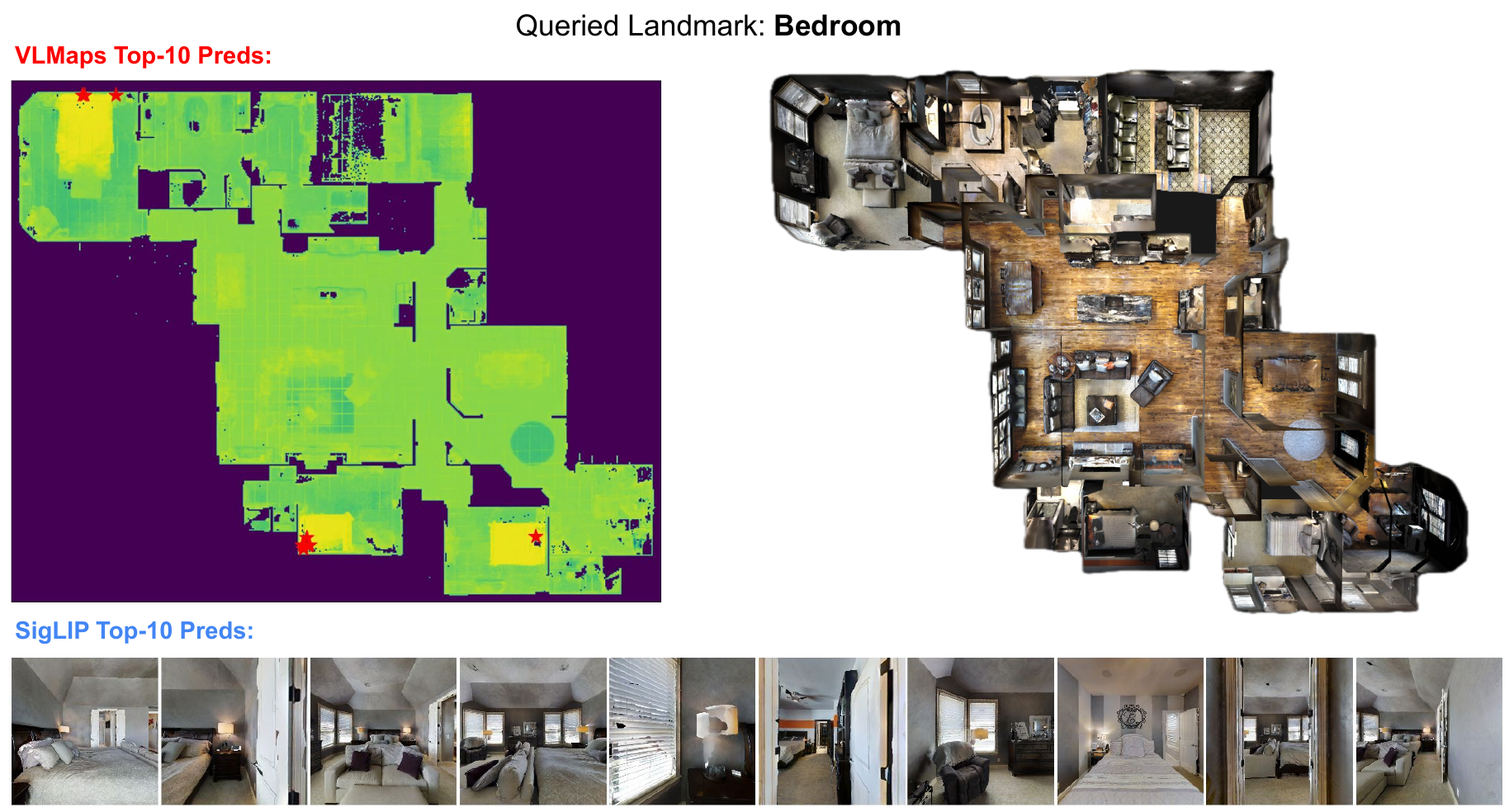}
        \caption{Bedroom}
        \label{fig:sub1}
    \end{subfigure}
    \hfill 
    \begin{subfigure}[t]{0.49\textwidth}
        \centering
        \includegraphics[width=\linewidth]{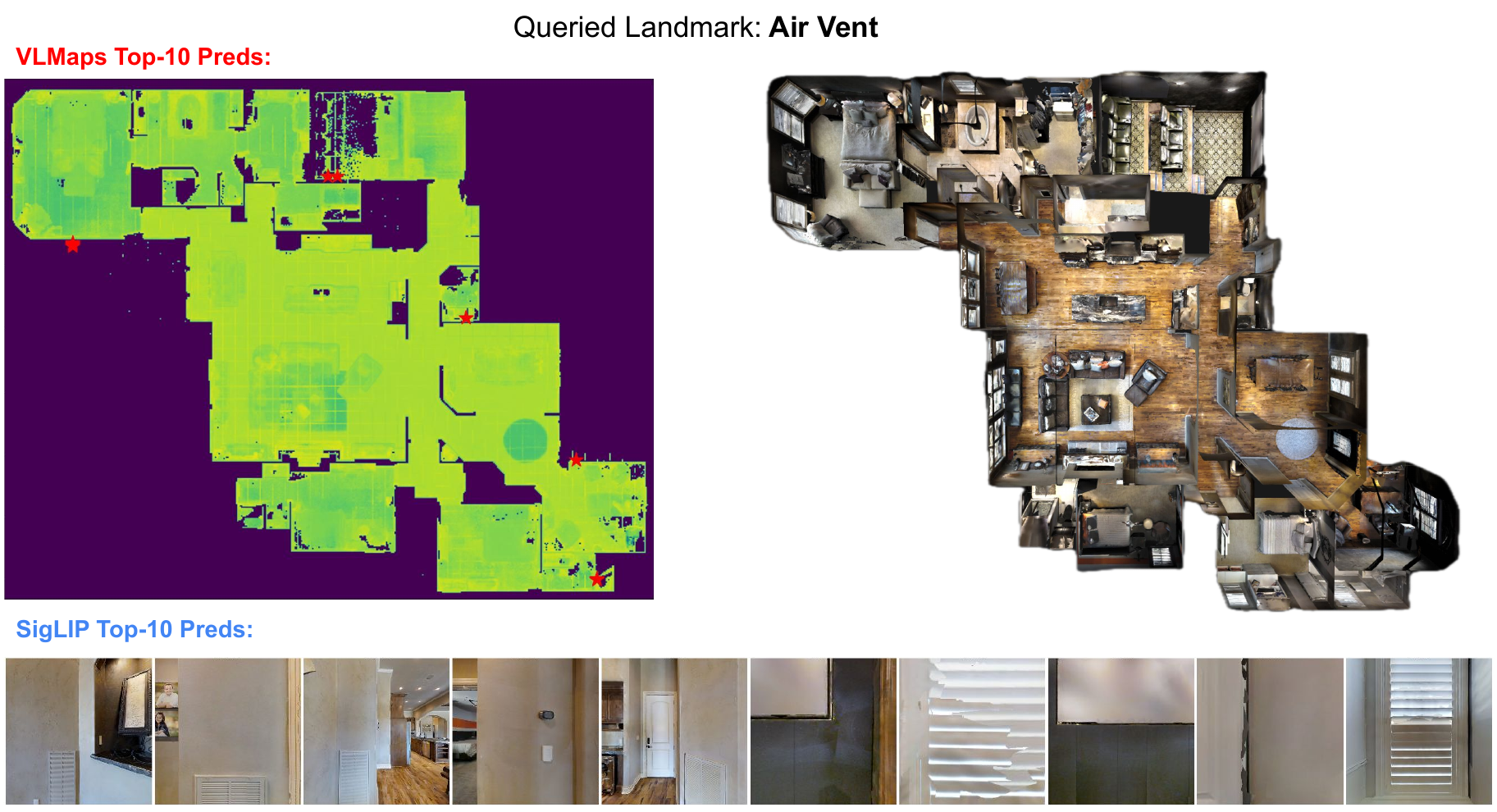}
        \caption{Air Vent}
        \label{fig:sub2}
    \end{subfigure}
    \caption{SigLIP vs. VLMaps Query Result for Last Landmark Indexing}
    \label{fig:siglipVSvlmaps}
\end{figure*}

We use the state-of-the-art vision language model SigLIP \cite{siglip} for goal/final landmark recognition, as shown in Figure \ref{fig:siglipVSvlmaps}. SigLIP training is similar to the CLIP model, replacing the contrastive loss with sigmoid binary prediction. The recognition is carried out by computing cosine similarity between panorama images and the textual description of the landmark. 
In order to compare the effectiveness of this choice with an open-vocabulary semantic map such as VLMaps \cite{vlmaps} that endows the occupancy map with CLIP embeddings, 
we ran the landmark localization experiment on all 127 landmarks and reported the mean Precision@10 in Table \ref{tab:vlmapsSiglipRetrievalResults}. The superiority of our approach stems from recognizing the landmarks in the panoramic views and replacing CLIP \cite{clip} with SigLIP \cite{siglip}, instead of using open-vocabulary semantic occupancy maps.

\begin{table}[!h]
\centering
{\footnotesize
\begin{tabular}{lcc}
\toprule
Model & \# Landmarks & Precision@10 ($\%$)\\ \midrule
VLMaps \cite{vlmaps} w/ CLIP \cite{clip} &  127 & 34.4 \\ \midrule
\rowcolor{verylightgreen} \textbf{Ours} w/ SigLIP \cite{siglip} & 127 & \textbf{70.0} \\
\bottomrule \\
\end{tabular}}
\caption{SigLIP vs. VLMaps Quantitative Results for Last Landmark Indexing}
\label{tab:vlmapsSiglipRetrievalResults}
\vspace*{-1em}
\end{table}


\begin{figure*}[!h]
  \centering
  \includegraphics[scale=0.45]{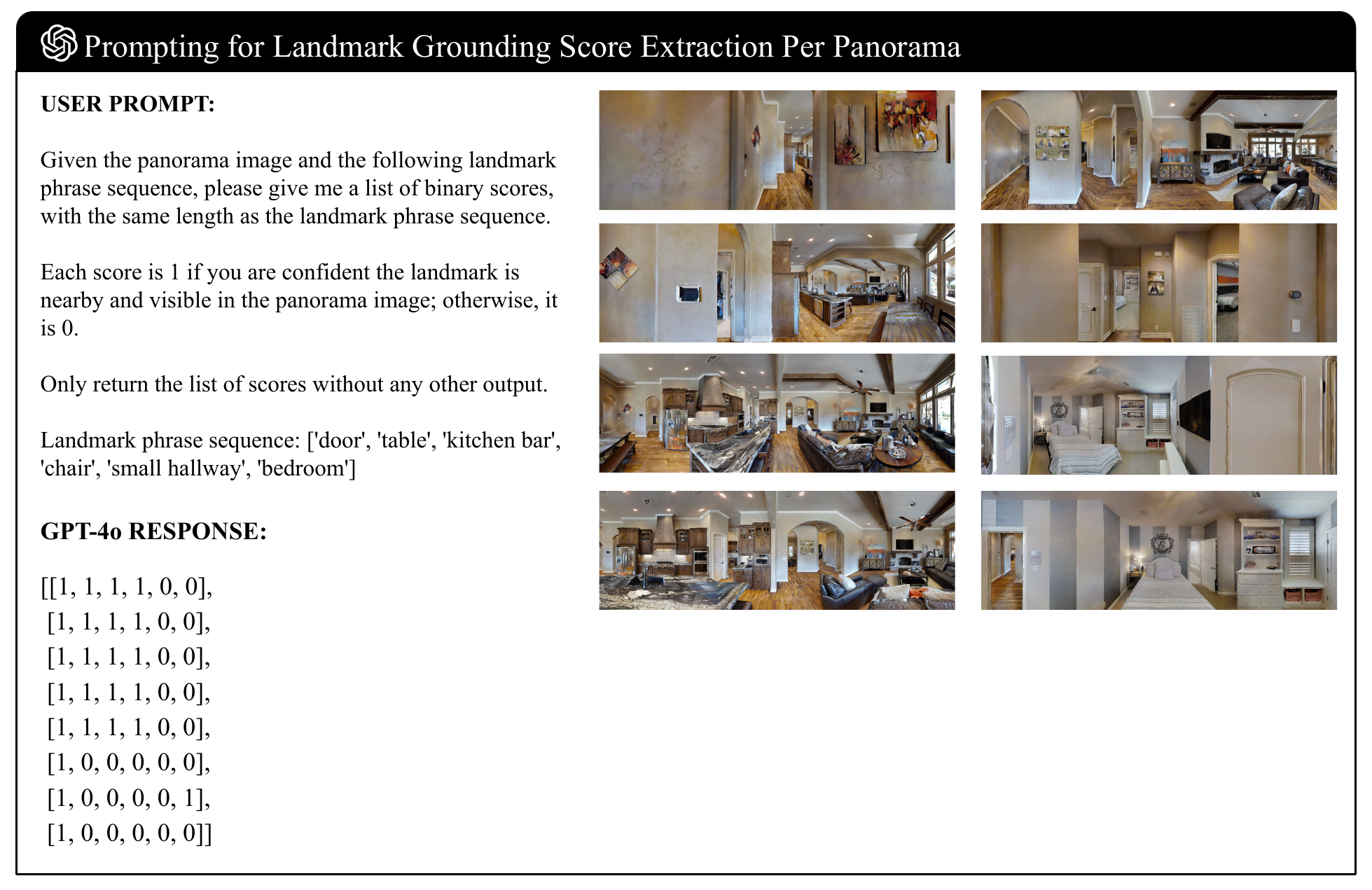}
  \caption{GPT-4o Landmark Grounding Score Extraction}
  \label{fig:gpt4o_grounding_score}
\end{figure*}

\begin{figure*}[!h]
  \centering
  \includegraphics[width=\linewidth]{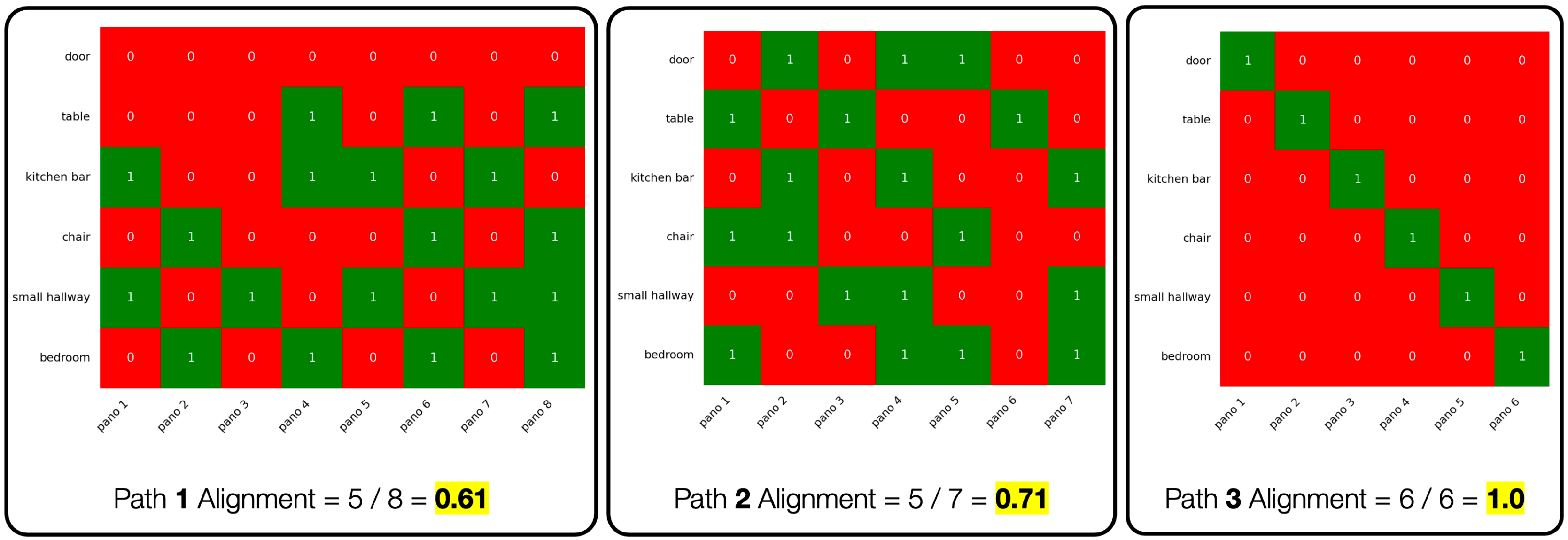}
  \caption{Sequence Alignment for Path Ranking (Pano2Land)}
  \label{fig:seqalignment}
\end{figure*}

Given the top-k goal locations, we compute the BFS shortest path from the starting pose to the goal nodes, obtaining k paths hypotheses. In the next stage, we quantify the alignment of the instruction with each of the paths and select the one with the highest alignment score. We introduce two approaches for path-instruction alignment and ranking. In \textsc{Approach I}, we formulate this problem as a sequence-to-sequence alignment, where the sequence of panoramas is  $X = \left[X_{0}, X_{1}, ..., X_{p}\right]$, and the sequence of landmark phrases $Y = \left[Y_{0}, Y_{1}, ..., Y_{l}\right]$, as shown in Figure \ref{fig:seqalignment}. Considering $X \times Y$ as a matrix $A$, where $A_{ij}$ is the binary grounding scores
of landmark being present in the panorama associated with the waypoint. We use the state-of-the-art VLM, GPT-4o, in our case, as shown in Figure \ref{fig:gpt4o_grounding_score} to obtain these scores.
We first discard the path hypotheses where the number of nodes is smaller than the number of landmarks. Then, given the $A$ matrix 
we compute for each path the normalized alignment score using Dynamic Programming (DP) formulation similar to the Longest Common Subsequence (LCS) problem, named Pano2Land described in Algorithm \ref{alg:algo}. 
%

Figure \ref{fig:seqalignment} shows the alignment matrix $A$  for three path hypotheses, comprised of 8, 7, and 6 nodes. The left example yields a score of 5/8, corresponding to 5 of the landmark names being successfully grounded in the right order in 8 consecutive panoramas. The middle example yields a score of 5 by grounding landmarks in panoramas 2, 3, 4, 5, and 7, where the final score would be $5/7 = 0.71$. The right example demonstrates the perfect way of aligning all the panoramas to the corresponding landmarks without skipping, yielding the top score of $6/6 = 1$.

\begin{figure*}[!h]
  \centering
  \includegraphics[width=\linewidth]{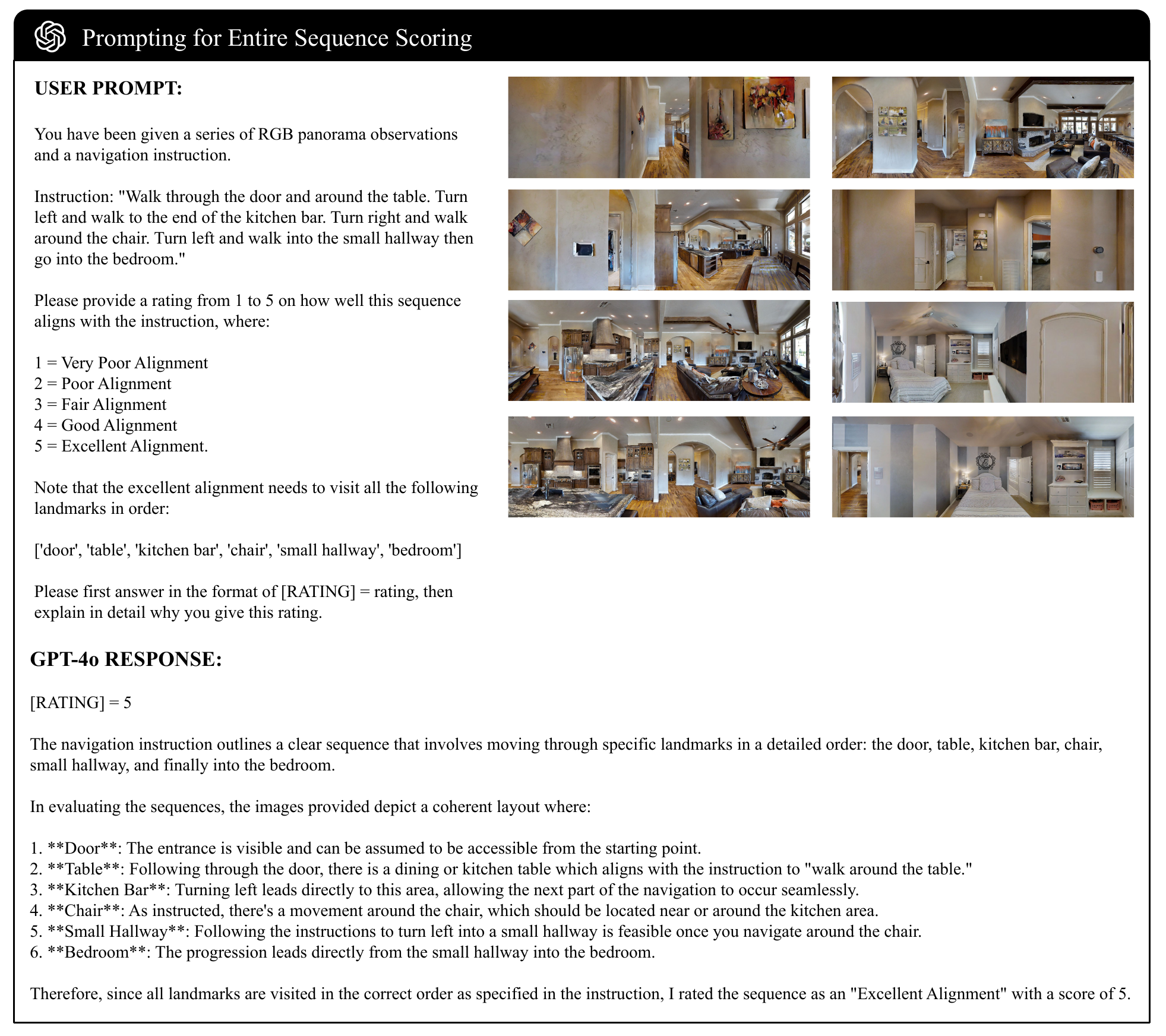}
  \caption{GPT-4o Full Prompting for Entire Sequence Scoring}
  \label{fig:gpt4o_full_prompting}
\end{figure*}

Alternatively, we introduce \textsc{Approach II} for path ranking by prompting GPT-4o to rate each path on a scale of 1 to 5 given the sequence of panoramas in order, original natural language instruction, and the extracted sequence of landmark phrases, as shown in Figure \ref{fig:gpt4o_full_prompting}. This approach bypasses the individual landmark grounding stage and alignment score computation done by \textsc{Pano2Land} algorithm. 
The performance of this approach is slightly worse than \textsc{Approach I}. Furthermore, the results are less interpretable since the internal ranking mechanism of GPT-4o is unknown. 

\begin{table*}[!h]
\centering
{
\begin{tabular}{lc|c|c|cc|ccc}
\toprule
& \multicolumn{1}{c}{} & \multicolumn{1}{c}{\textsc{Num}} & \multicolumn{1}{c}{\textsc{Hypo Path Gen}} & \multicolumn{2}{c}{\textsc{Approach \textbf{I}}} & \multicolumn{2}{c}{\textsc{Approach \textbf{II}}} \\
& & Episodes & Accuracy ($\%$) & nDTW ($\%$)  & Accuracy ($\%$)  & nDTW ($\%$) & Accuracy ($\%$) \\ \midrule
\texttt{8WUmhLawc2A} & & 21 & 66.7 & 88.68$\pm$0.0 & 57.10$\pm$0.0 & 87.34$\pm$0.52 & 52.38$\pm$6.73 \\
\texttt{JeFG25nYj2p} & & 21 & 61.9 & 87.51$\pm$0.13 & 52.38$\pm$0.0 & 88.92$\pm$0.64 & 49.20$\pm$5.93\\
\texttt{mJXqzFtmKg4} & & 21 & 66.7 & 91.21$\pm$0.07 & 66.70$\pm$0.0 & 90.08$\pm$0.21 & 57.14$\pm$0.0 \\
\texttt{r1Q1Z4BcV1o} & & 21 & 57.1 & 87.96$\pm$0.30 & 49.20$\pm$2.24  & 88.69$\pm$0.42 & 39.68$\pm$2.24 \\
\texttt{sT4fr6TAbpF} & & 21 & 76.2 & 89.25$\pm$1.05 & 61.90$\pm$3.88   & 86.68$\pm$0.34 & 52.38$\pm$6.73\\
\midrule
\textbf{\texttt{Average}} & & 105 & \textbf{65.72}$\pm$6.33 & \textbf{88.92}$\pm$1.28 & \textbf{57.45}$\pm$6.31 & 88.34$\pm$1.20  & 50.15$\pm$5.81 \\
\bottomrule \\
\end{tabular}}
\caption{Full Pipeline Quantitative Results}
\label{tab:vln_final_results}
\end{table*}

Finally, for each approach's output, we compute the normalized dynamic-time warping (nDTW) metric between the ground truth and the best-aligned path to measure the path fidelity; nDTW is more aligned with our task objective compared to the Success Rate (SR), which only considers an episode to be successful if the agent's last position is within 3 meters of the ground-truth goal and it does not explicitly consider the intermediate alignments with the landmarks that were supposed to be visited in order by the agent~\cite{vlnpathfidelity}.

\begin{algorithm}
\footnotesize
\caption{- \textsc{Pano2Land} algorithm for calculating path alignment using grounding scores, similar to DP/LCS.}
\begin{algorithmic}[1]
\Require Binary grounding matrix $\mathbf{M} \in \{0,1\}^{R \times C}$
\Ensure Alignment score $S$

\State $R \gets$ number of rows (landmarks) in $\mathbf{M}$
\State $C \gets$ number of columns (panoramas) in $\mathbf{M}$
\State Initialize matrix $\mathbf{dp} \in {N}_0^{(R+1) \times (C+1)}$ with zeros

\For{$r \gets 1$ to $R+1$}
    \For{$c \gets 1$ to $C+1$}
        \If{$M_{r,c} == 1$}
            \State $\mathbf{dp}[r][c] \gets (\mathbf{dp}[r-1][c-1]) + 1$
        \Else
            \State $\mathbf{dp}[r][c] \gets \max\left(\mathbf{dp}[r-1][c],\; \mathbf{dp}[r][c-1] \right)$
        \EndIf
    \EndFor
\EndFor

\State $S \gets \mathbf{dp}[R][C]$
\State \Return $S$
\label{alg:algo}
\end{algorithmic}
\end{algorithm}




       

In Table \ref{tab:vln_final_results}, 
\textsc{Hypo Path Gen} accuracy indicates the fraction of episodes where the ground-truth path or a highly similar one is among the selected path hypotheses. There might be multiple reasons why the correct path couldn't be retrieved, including but not limited to (1) not being able to ground the last landmark, (2) encountering a dramatically different landmark that has been part of the train samples, (3) a highly-frequent last landmark which exists in multiple locations (e.g., door) where the ground-truth landmark location may not fall into the top-3 retrieved ones, etc.

The \textsc{Approach I} nDTW shows the average nDTW score for all 21 episodes in each environment.
If the nDTW is above 87\%, we consider the path successful, where this threshold is based on our empirical analysis. We applied the same evaluation metrics to  \textsc{Approach II} 
Both numbers are higher in the \textsc{Approach I}. While in \textsc{Approach II} the full natural language instruction, including the action phrases (e.g. turn left) is given to GPT-4o, we hypothesize that the model has trouble grounding actions and/or landmarks and \textsc{Approach I} benefits from explicit decomposition of different stages. Since there were cases in which multiple top grounding scores or GPT-4o rating scores would exist, we repeated the process of picking the path with the highest score randomly up to 3 times and reported the mean and standard deviation.

\section{Limitations}
There are specific limitations to our approach that we'd like to elaborate on. Firstly, our approach only works in the previously explored environments, given the topological map. Secondly, it only works in cases where the natural language instruction is landmarks-based and is not heavily based on spatial and temporal phrases, action phrases, and absolute metric distances. Since our pipeline is modular and not trained end-to-end, drawbacks of each module, especially the early stages of the LLM landmark extraction and VLM retrieval, propagate the errors to later stages of \textsc{Pano2Land} alignment or GPT-4o ranking. The quality of the path hypotheses eventually determines the upper bound on the ranking computed by GPT-4o or any other VLM being used.

\section{Conclusion}
In this work, we introduced a modular approach for the vision-and-language navigation (VLN) task based on the R2R-Matterport3D dataset \cite{r2r, chang2017matterport3d} within the Meta Habitat Simulator \cite{habitat19iccv, Yadav2022HabitatMatterport3S}. Our approach assumes that the agent has built a topological map in the exploration stage. We then use LLM to extract the sequence of landmarks the agent needs to visit, retrieve the top-k goal locations, and rank the path hypotheses to select the one with the highest alignment with the natural language instructions as the final answer. For the task, the approach demonstrates the superiority of the topological map with per-node panoramas to an open-vocabulary semantic occupancy map for land-mark grounding and goal retrieval. The overall performance on this benchmark is mainly affected by the zero-shot capabilities of VLM's to ground special landmark names in the panoramas. Future improvements can be 
achieved by fine-tuning the existing VLMs on navigation tasks and deploying the agent in previously unseen environments by seamlessly integrating the exploration and navigation part.

{
    \small
    \bibliographystyle{ieeenat_fullname}
    \bibliography{main}
}


\end{document}